\documentclass[journal]{IEEEtran}

\makeatletter
\def\endthebibliography{%
	\def\@noitemerr{\@latex@warning{Empty `thebibliography' environment}}%
	\endlist
}
\makeatother

\usepackage[noadjust]{cite}
\ifCLASSINFOpdf
  \usepackage[pdftex]{graphicx}
  \graphicspath{{pdf/}{jpg/}{png/}}
\else
\fi
\usepackage{amsmath}
\usepackage{cases}

\usepackage{array}

\usepackage{stfloats}
\fnbelowfloat
\usepackage{gensymb} % for \degree, etc.
\usepackage{xspace} % to fix \degree eating a space
\AtBeginDocument{%
	\let\OLDdegree\degree%
	\renewcommand{\degree}{\OLDdegree{}\xspace}%
}

\usepackage{xcolor} % for inkscape figures

\usepackage{footnote}
\makesavenoteenv{tabular}
\makesavenoteenv{table}
\usepackage{multirow}
\usepackage{citesort}

\newif\ifLPRfmtDRAFT
\newif\ifLPRfmtHIGHLIGHTS
\newif\ifLPRfmtNOAUTH

\newif\ifNO
\newif\ifYES
\YEStrue

\LPRfmtHIGHLIGHTStrue
\LPRfmtDRAFTtrue

\ifLPRfmtDRAFT

	\usepackage[colorinlistoftodos]{todonotes}
	\usepackage{mdframed}
	\usepackage{scrtime}
	\usepackage{soul}

	\mdfdefinestyle{notesstyle}{%
		linecolor=red,frametitlerule=true,
		backgroundcolor=red!15}
	\mdfdefinestyle{outlinestyle}{%
		linecolor=green,frametitlerule=true,
		backgroundcolor=green!15}
	
	\presetkeys{todonotes}{color=blue!15, linecolor=red!90, size=\footnotesize, figwidth=3.4in}{}
	\makeatletter
	\if@todonotes@disabled
		\newcommand{\hlfix}[2]{#1}
	\else
		\newcommand{\hlfix}[2]{\texthl{#1}\todo{#2}{}}
	\fi
	\makeatother
\else
	\makeatletter
	\newcommand{\hlfix}[2]{{}}
	\newcommand{\todo}[1]{{}}
	\makeatother
\fi

\ifLPRfmtHIGHLIGHTS
	\colorlet{editColorBlue}{blue!100!}
	\colorlet{editColorRed}{red!100!}
	\definecolor{editColorGreen}{RGB}{0,127,0}
	\definecolor{editColorRedRed}{RGB}{192,0,128}
	\definecolor{editColorBlueBlue}{RGB}{0,96,192}
	
	\newcommand{\editB}[1]{{\color{editColorBlue}#1}}

\else
	\newcommand{\editB}[1]{#1}

\fi

\begin{document}
\title{End-to-End Domain Adaptive Attention Network for Cross-Domain Person Re-Identification}

%\author{Michael~Shell,~\IEEEmembership{Member,~IEEE,}
%        John~Doe,~\IEEEmembership{Fellow,~OSA,}
%        and~Jane~Doe,~\IEEEmembership{Life~Fellow,~IEEE}% <-this % stops a space
\ifLPRfmtNOAUTH
\author{}% None of that if we're hiding authorship information.
\else
\author{Amena~Khatun,~\IEEEmembership{Student~Member,~IEEE,}
		Simon~Denman,~\IEEEmembership{Member,~IEEE,}
		Sridha~Sridharan,~\IEEEmembership{Life Senior~Member,~IEEE,}
		Clinton~Fookes,~\IEEEmembership{Senior~Member,~IEEE}% <-this % stops a space
%\thanks{M. Shell was with the Department
%of Electrical and Computer Engineering, Georgia Institute of Technology, Atlanta,
%\thanks{L. Renaud, Y. Labmate, and Y. Professor are with the School of Electrical Engineering and Computer Science, My University, City, ST, ZIPCODE USA (e-mail: l_renaud@univ.edu; y_labmate@univ.edu; y_professor@univ.edu).}% <-this % stops a space
%\thanks{Manuscript received April 19, 2005; revised August 26, 2015.}%
}
\fi

\maketitle

\begin{abstract}
Person re-identification (re-ID) remains challenging in a real-world scenario, as it requires a trained network to generalise to totally unseen target data in the presence of variations across domains. Recently, generative adversarial models have been widely adopted to enhance the diversity of training data. These approaches, however, often fail to generalise to other domains, as existing generative person re-identification models have a disconnect between the generative component and the discriminative feature learning stage. To address the on-going challenges regarding model generalisation, we propose an end-to-end domain adaptive attention network to jointly translate images between domains and learn discriminative re-id features in a single framework. To address the domain gap challenge, we introduce an attention module for image translation from source to target domains without affecting the identity of a person. More specifically, attention is directed to the background instead of the entire image of the person, ensuring identifying characteristics of the subject are preserved. The proposed joint learning network results in a significant performance improvement over state-of-the-art methods on several benchmark datasets.%
%\ifLPRfmtDRAFT\footnote{Compiled on \today\: at \thistime}\fi
\end{abstract}%

\editB{%
}

% \fi
\IEEEpeerreviewmaketitle

\section{Introduction}\label{sec:10-intro}%
\IEEEPARstart{P}{erson} re-identification (re-ID) is becoming increasingly important in security and surveillance, and aims to find a target person from a large gallery set captured by different cameras. As such, it can be viewed as a domain adaptation task. Although person re-ID is a widely investigated research area, it is still challenging as images of a person often undergo intensive changes in illumination, background, pose, and viewpoint due to the domain shift  between two cameras, leading to a severe drop in performance in a real-world scenario where the domain of the target images has no overlap with the domain of the gallery images. The domain shift is illustrated in Figure 1.

To address the challenge of domain variation, re-ID researchers adopted invariant feature representation methods such as XQDA \cite{XQDA}, KISSME \cite{KISSME} or deep learning based methods such as IDE \cite{Zheng2016PersonRP}, and Siamese \cite{DBLP:journals/corr/VariorHW16} or Triplet \cite{Cheng_2016_CVPR} networks to help deal with different cameras. However, while these methods have helped relax the closed-world assumptions of previous methods, they still suffer from performance degradation when confronted with a real-world scenario where conditions in the target images are totally unseen. 
To address the domain gap issue in other tasks such as object recognition, domain adaptation researchers \cite{coral,dcoral,G2A,DupGAN} have adopted unsupervised domain adaptation (UDA) techniques that consider the labeled source data and unlabeled target data, and thus assume that the source and target domains share the same label, i.e. the same set of classes. For example, in the object classification task, both domains may have the class `backpack'. However, this scheme does not work for person re-ID as every person has a different identity, thus belonging to a different class. Therefore, UDA approaches cannot be applied directly to re-ID to solve the domain gap issue.

\begin{figure}
\begin{center}
\includegraphics[width=1.0\linewidth]{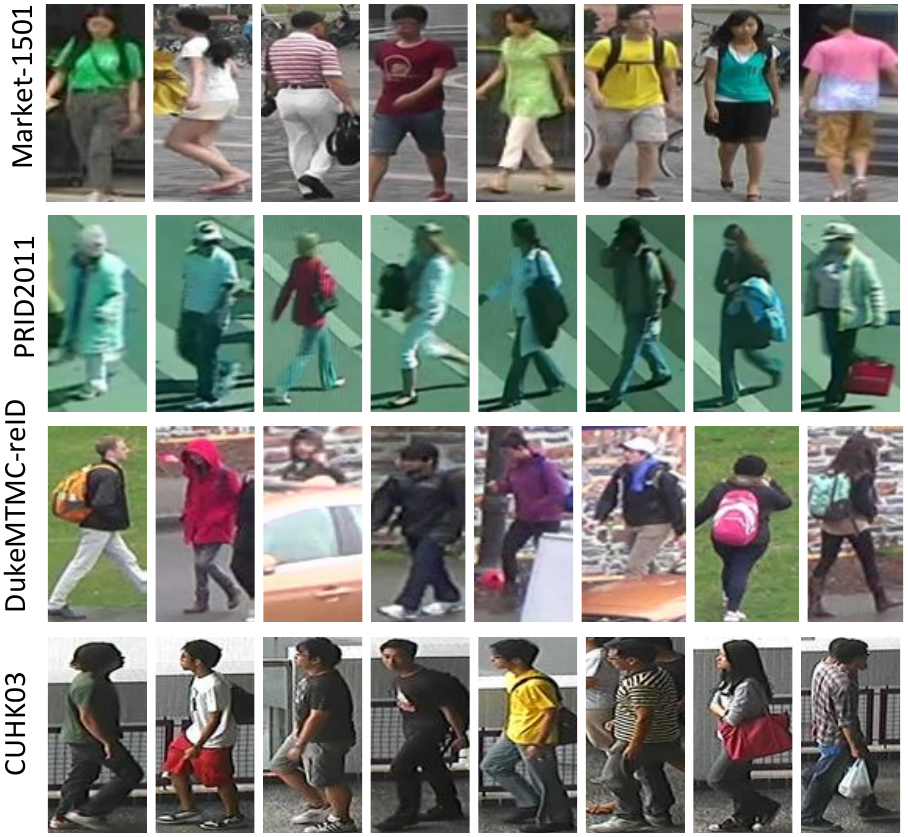}
\end{center}
\caption{Domain variance comparison of Market1501, PRID2011, DukeMTMC-reID and CUHK03 benchmarks, showing significant changes in illumination, pose, background and camera viewpoint.}
\label{fig:Dataset_comparison}
\end{figure}

Recently, generative adversarial models are becoming popular to address the domain gap by introducing more variations into the training data. Current generative re-ID methods adopt either CycleGAN \cite{CycleGAN2017} or StarGAN \cite{8579014} for image translation from camera-to-camera or domain-to-domain, and achieve significant performance gains. CycleGAN \cite{CycleGAN2017} translates images from one domain to another using the cycle-consistency loss without any paired input data. Inspired by the success of CycleGAN, many re-ID researchers \cite{Wei2018PersonTG,Bak_2018_ECCV,image-image18,zhong2018camera} applied cycle-consistency loss for image translation to reduce the domain gap. However, CycleGAN is not able to translate specific parts of an image, and instead alters the entire image, interfering with the appearance and attribute of a person. Unintentionally, CycleGAN based re-ID methods affect the foreground of images (i.e. the person's appearance) instead of just the background. Thus, it hurts person re-ID performance as we aim to transform only the color and structure of the background from one domain to another, rather than altering the entire image. As such, despite the use of methods such as cycle-GAN, existing re-ID models do not generalise well when trained on one domain and tested on another due to dramatic variations in appearance between cameras, and this leads to a severe drop in performance. For example, when the PRID2011 dataset is evaluated using a model trained on CUHK03 with no adaptation, the rank-1 accuracy is only 9.5\%. Moreover, generative adversarial network (GAN) based methods require two stages to perform re-ID. In the first stage, a GAN is used for image domain translation. The generated synthetic images are then used to train the re-ID model in the second stage which restricts to gain the benefits from the generated samples as the optimisation target of the image generation model cannot be well-matched with the re-ID.

Motivated by the above observations, and to address the challenges posed by the domain gap, we propose an end-to-end domain adaptive attention network (EDAAN) for re-ID, to pay attention to specific regions during image translation and better preserve identity information. The proposed attention network minimises the divergence only between the specific region for the source and target domain, in contrast to the existing methods which minimise the discrepancy between the distribution of the whole image from the source and target domains. We also demonstrate that re-ID performance can be further improved if the image generation module and the discriminative re-ID module are trained jointly, allowing them to support each other such that during image translation between domains, the generative module gains knowledge of a person's identity via the discriminative module, and the discriminative module can learn the appearance of a person in the target domain from the generative module. Thus, the proposed framework not only focuses on preserving the foreground of a person during image translation, but jointly optimises the generative and re-ID network to leverage advantages from each.

Within the proposed framework, we adopt our earlier proposed quartet loss \cite{Amena} to boost the re-ID performance as this loss minimises intra-class distance more than the inter-class distance in the feature space. In \cite{Amena}, we argue that the quartet loss performs better than the traditional triplet loss \cite{AAAI1714313} as the quartet loss minimises the intra-class variation over the inter-class variation, regardless of whether the probe image belongs to the same person or not; while the triplet loss pushes images of the same identity close to each other only when the probe images come from the same identity, which is not practical in a real-world application. The major contributions of this paper can be summarised as follows:
%\vspace{-2mm}
\begin{itemize}
    \item We propose a novel attention network to preserve the foreground of a person and keep a person's identity and appearance constant while adapting the background from another domain. 
    \item We minimise the domain gap by successfully transforming one domain style to another domain and achieve a significant gain in performance in cross-domain evaluations.
    \item We propose an end-to-end network to jointly optimise an attention based generative network and discriminative re-ID network as a unified system, allowing each network to leverage knowledge from the other.
\end{itemize}

\section{Related Work}

\subsection{Deep Re-ID Feature Learning}
A large volume of re-ID research \cite{6909421,7299016,7780513,DBLP:journals/corr/VariorSLXW16,DBLP:journals/corr/VariorHW16,6976727,6909576,Ding:2015:DFL:2796563.2796623,Cheng_2016_CVPR,DBLP:journals/corr/ChenCZH17,8099845,Zhao_2017_ICCV,Zhang2017AlignedReIDSH,AAAI1714313,Amena,98a1e05749b24099a51dcf3c22daefd9,Li_2017_CVPR,Yang2019PatchBasedDF,zhao2017spindle,Fan:2018:UPR:3282485.3243316,Lin2019ABC,DECAMEL,8976262,8922622,8732420} adopts deep learning approaches as they combine feature extraction and metric learning in a single framework. Some methods focus on verification \cite{Cheng_2016_CVPR} or identification \cite{DBLP:journals/corr/VariorHW16,7780513} losses while others combine both approaches \cite{AAAI1714313,Amena,Khatun_2020_WACV}, and adopt the siamese \cite{6909421,7299016,7780513,DBLP:journals/corr/VariorSLXW16,DBLP:journals/corr/VariorHW16,6976727}  or triplet loss \cite{6909576,Ding:2015:DFL:2796563.2796623,Cheng_2016_CVPR,AAAI1714313}. The authors in \cite{AlignGAN} reduces cross-modality discrepancies by  generating cross-modality images and adopt the triplet loss for verification. In \cite{DBLP:journals/corr/ChenCZH17}, the triplet loss was improved by a new loss function which used four input images \cite{DBLP:journals/corr/ChenCZH17,Amena} instead of three during training. A number of recent approaches have used part based methods \cite{98a1e05749b24099a51dcf3c22daefd9,Li_2017_CVPR,Yang2019PatchBasedDF,zhao2017spindle} with the aim of overcoming occlusions, and handling partial observations.
Other methods \cite{Fan:2018:UPR:3282485.3243316,Lin2019ABC,DECAMEL} are focused on clustering or transferring the knowledge from a labeled source dataset to unlabelled data using pseudo labels. However, different identities may have the same pseudo label which can make it hard for the model to distinguish similar people. 

\subsection{Unsupervised Domain Adaptation}

Another active research path to address the domain gap problem is unsupervised domain adaptation. Some research utilises unsupervised domain adaptation to the reduce discrepancy between domains using Maximum Mean Discrepancy
(MMD) \cite{DBLP:conf/icml/LongC0J15} or Correlation alignment \cite{sun2016deep}. As the target domain does not contain labels, all UDA approaches \cite{DBLP:conf/icml/LongC0J15,sun2016deep,coral,dcoral,G2A,DupGAN,RAHMAN2019107124} assume that the target domain shares the same labels as the source domain where the classes of
the source and target domains are entirely identical, i.e. a closed-set scenario. In \cite{Li_2018_CVPR_Workshops}, UDA is performed for person re-ID to minimise the domain gap challenge by mapping existing labeled source data to unlabeled target data. However, they also assume that the source domain and target domain will have the same number of classes which is impractical in a real-world person re-ID scenario. Hence, for the open set problem of person re-ID, the approaches mentioned above are unsuited for the problem of unsupervised domain adaptation in person re-ID, where each person belongs to a different class. 
\begin{figure*}
\begin{center}
\includegraphics[width=1.0\linewidth]{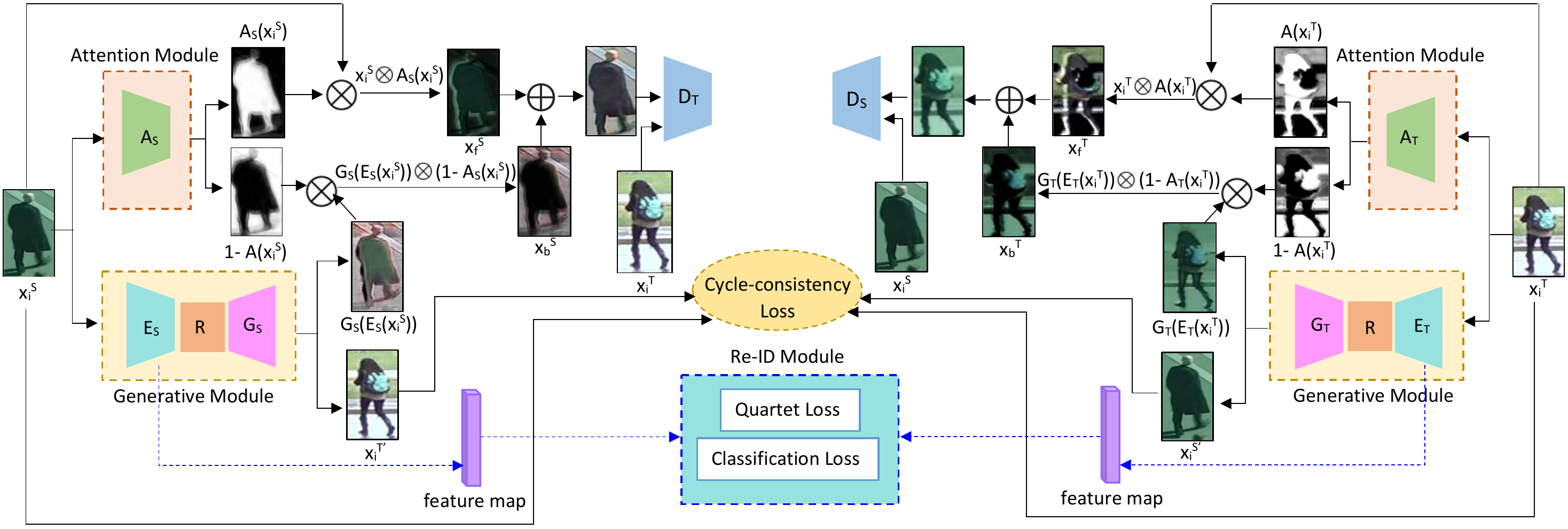}
\end{center}
\caption{An illustration of the proposed EDAAN for image translation from one domain to another. A, E, G, and D represent the attention network, the generative module for image translation and the discriminator. $A(x_i)$ and  $1 - A(x_i)$ represent the attention maps produced by the attention network. $\otimes$ denotes element-wise multiplication.}
\vspace{-4mm}
\label{fig:architecture}
\end{figure*}

In recent person re-ID studies, CycleGAN based UDA approaches have been widely investigated to enhance input variations and improve cross-domain performance. In \cite{zhong2018camera}, the authors proposed to use CycleGAN to transfer the styles between cameras with label smoothing regularization for camera-style adaptation. In \cite{Wei2018PersonTG}, cycle-consistency loss is used to to bridge the domain gap and an identity loss is used to preserve the identity. The authors in \cite{Bak_2018_ECCV} solved the domain shift issue in three stages: an illumination module to render pedestrians in different illumination conditions, a domain translation CycleGAN module to transform images from one domain to another and a feature learning module for re-ID. SPGAN is proposed in \cite{image-image18} to preserve the self-similarity of translated image and domain dissimilarity of the source domain's real image, which is very close to the approach of \cite{zhong2018camera}. However, these methods adopted CycleGAN and thus distorts the appearance of a person by transferring the entire style of the source domain to the target domain, and leads to a performance drop. In addition, all these methods perform re-ID in two disjoint phases. The first phase is utilised for image generation while the second addresses on feature learning, restricting the models ability to learn from each other.
\vspace{-2mm}
\subsection{Comparison of proposed approach to previous works}
In this work, we focus on translating the color and structure of the background from one domain to another, without affecting the appearance of a person (i.e. what makes them unique). To achieve this aim, we produce attention maps using an attention network to tell the network where to focus during image translation. 

Unsupervised attention learning is also studied in \cite{mnih2014recurrent,zhang2018self,yang2017lr,NIPS2016_6194,chen2018attention}. In \cite{mnih2014recurrent}, a square region of the image is used for visual attention while \cite{yang2017lr} generates the foreground and background separately and recursively to produce a natural image. However, these methods are not applied for image translation. A self-attention network is proposed in \cite{zhang2018self} to generate globally realistic images. Attention maps are used in \cite{NIPS2016_6194} for video generation and in \cite{chen2018attention} for image translation, however, \cite{chen2018attention} modifies the foreground only and their work is limited to image generation. In contrast, we aim to adapt the background from another domain and preserve the appearance of a person and perform re-identification in an end-to-end manner. Our proposed method incorporates image translation and feature learning in a unified architecture. Since image translation can receive feedback from the re-ID feature learning and reinforce itself, our method better leverages the two components to maximise performance.
\vspace{-2mm}
\section{Our approach}

As illustrated in Figure \ref{fig:architecture}, our EDAAN (end-to-end domain adaptive attention network) tightly couples the generative and discriminative networks for image translation and re-ID learning within a single architecture. EDAAN consists of an attention-based generative module and a discriminative re-ID module. The attention-based generator is responsible for creating an attention mask to predict the region of interest and translating image between domains; while the discriminative component uses the information from the generative network to learn how to identify people.
%\vspace{-6mm}
\subsection{Domain Adaptive Attention Module}
%\vspace{-2mm}
\subsubsection{Problem Formulation}
We denote the real images from the source domain, S, and target domain, T, as $x^{S}_{i}$ and $x^{T}_{i}$ respectively. The attention network $A_S$ is built within the image generation network to select the background of an image to translate during image translation without impacting the foreground. The generated foreground and background attention maps are denoted as $A_S(x_i^S)$ and $1- A_S(x_i^S)$ respectively for the source domain, $S$. A higher score is allocated by the attention network to the background that ranges from 0 to 1 per-pixel while ignoring the foreground of a person's image. The input image $x_i^S$ is passed through the image generator $G_S(E_S)$ to transform the source domain image to the style of target domain. The learned background mask from the attention network is then applied to the transformed image $G_S(E_S(x_i^S))$ by a layered operation, i.e. an element-wise product, to generate a new image $x_b^S$ which adapts the background of the target domain. Another layered operation is performed between the real image and the foreground mask which generates another image $x_f^S$ to preserve the foreground of the real image. The images  $x_b^S$ and $x_f^S$ are then concatenated to produce the final image. A similar attention network $A_T$ is used for the target domain.

\subsubsection{Attention-Guided Image Translation Module}
The input image $x_i^S$ from the source domain $S$ is first passed through the attention module to produce two attention maps: the foreground mask, $A_S(x_i^S)$ and the background mask, $1- A_S(x_i^S)$. The same input image is fed to the generator module $G_S(E_S(x_i^S)): S \rightarrow T(x_i^S)$ to translate the source domain's image according to the background of the target domain images. Hence, the final mapping function for the source domain image $x_i^S$ can be represented by,
\begin{multline}
G(x_i^S)=\underbrace{(1\! - \!A_S(x_i^S))\!\odot\! G_S(E_S(x_i^S)): S\! \rightarrow \!T(x_i^S)}_\text{background}\\+ \underbrace{A_S(x_i^S)\!\odot\! x_i^S}_\text{foreground},
\end{multline}

%\begin{multline}
%G_S(E_S(x_i^S))=\underbrace{(1\! - \!A_S(x_i^S))\!\odot\! G_S(E_S(x_i^S)): S\! \rightarrow \!T(x_i^S)}_\text{background}\\+ \underbrace{A_S(x_i^S)\!\odot\! x_i^S}_\text{foreground},
%\end{multline}
%\begin{align}
%G_S(E_S(x_i^S))=\underbrace{(1 - \phi_A(x_i^A))\odot G_A: A  \rightarrow B(x_i^A)}_\text{background}
%+ \underbrace{\phi_A(x_i^A)\odot x_i^A}_\text{foreground}.
%\end{align}\
\vspace{2mm}
where $ \odot$ is an element-wise multiplication performed between the generated background masks and the domain translated images, and between the real input images and the forground masks.

Similarly, another mapping function is performed for the target domain,
%\begin{align}
%G_B(x_i^B) = \underbrace{\phi_B(x_i^B)}_\text{background}\odot G_B: B  \rightarrow A(x_i^B)
%+ \underbrace{(1 - \phi_B(x_i^B))}_\text{foreground}\odot x_i^B.
%\end{align}
\begin{multline}
F(x_i^T)=\underbrace{(1\! -\! A_T(x_i^T))\!\odot\! G_T(E_T(x_i^T)): T \!\rightarrow \!S(x_i^T)}_\text{background}\\+ \underbrace{A_T(x_i^T)\!\odot\! x_i^T}_\text{foreground}.
\end{multline}
%\begin{multline}
%G_T(E_T(x_i^T))=\underbrace{(1\! -\! A_T(x_i^T))\!\odot\! G_T(E_T(x_i^T)): T \!\rightarrow \!S(x_i^T)}_\text{background}\\+ \underbrace{A_T(x_i^T)\!\odot\! x_i^T}_\text{foreground}.
%\end{multline}
To learn the overall mapping, adversarial losses are used. For example, if we want to transfer the style of domain $S$ $\rightarrow$ $T$, the adversarial loss is given by,
\begin{align}
L_{GAN}^S(G, D_T, S, T) = _{x_i^T \sim P_{T}} [log D_T(x_i^T)]  \nonumber	\\
+ _{x_i^S \sim P_{S}} [log(1- D_S(G_S(E_S(x_i^S))))],
\label{eq:1}
\end{align}
where the mapping function is $G:S\rightarrow T$ and $D_T$ is the discriminator. The generator aims to generate images as the distribution of the target domain ($T$) while the discriminator, $D_T$, attempts to distinguish the generated images from real images. However, adversarial training requires paired training data, otherwise, infinitely many mappings will induce the same distribution over the output, and thus many input images will map to the same output image. To address this problem, we adopt the cycle consistency loss \cite{CycleGAN2017} to translate images $(x_i^S)$ from domain $S$ to domain $T$, and then translate it back from domain $T$ to domain $S$, and as such do not require paired training data. For example, two domains require two mapping functions which should be bijective. The cycle consistency loss can be expressed as,
\begin{align}
L_{cyc} (S, T) = _{x_i^S \sim P_{S}} [ || G_T(E_T({G_S(E_S}(x_i^S)))) - x_i^S||_1] \nonumber\\+ _{x_i^T \sim P_{T}} [ || G_S(E_S({G_T(E_T}(x_i^T)))) - x_i^T||_1]],
\label{eq:2}
\end{align}
%\vspace{2mm}
where $G_T(E_T({G_S(E_S}(x_i^S))))$ represents the reconstructed version of the real image $x_i^S$ from the source domain, and $G_S(E_S({G_T(E_T}(x_i^T))))$ is the reconstructed version of the real image, $x_i^T$ from the target domain.

The generated attention map of the input image also needs to be homogeneous with the attention map of the translated images, such that if person $a$ is translated with the background of person $b$, then the background of person $b$ should also be translated with the background of person $a$. Hence, an attention loss is used within the cycle-consistency loss which can be expressed as,
\begin{multline}
L_{attn}(A_S, A_T) = _{x_i^S \sim P_{S}} [ || A_S(x_i^S) - A_T(G(x_i^S)) ||_1]\\ + _{x_i^T \sim P_{T}} [ || A_T(x_i^T) - A_S(F(x_i^T)) ||_1],
\label{eq:5}
\end{multline}

%\begin{multline}
%L_{attn}\!(A_S, \!A_T) \!= \!_{x_i^S \sim P_{S}} [ || A_S(x_i^S) - \!A_T\!(\!G_T(\!E_T(\!{G_S(\!E_S}\\(x_i^S)\!)\!)\!)\!) ||_1] + _{x_i^T \sim P_{T}} [ || A_T(x_i^T) - A_S({G_S}({E_S}({G_T(E_T}(x_i^T)\!)\!)\!)\!) ||_1],
%\label{eq:5}
%\end{multline}

%\begin{multline}
%L_{attn}\! = _{x_i^S \sim P_{S}}\! [ || A_S(x_i^S) -\! A_T(\!G_T(\!E_T(\!{G_S(\!E_S}(x_i^S)\!)\!)\!)\!) ||_1] \\+\! _{x_i^T \sim P_{T}} [ || A_T(x_i^T)\! - \!A_S(\!{G_S}(\!{E_S}(\!{G_T(\!E_T}(\!x_i^T)\!)\!)\!)\!) ||_1],
%\label{eq:5}
%\end{multline}
where $A_S(x_i^S)$ is the attention map of the input image from the source domain and $A_T(G(x_i^S))$ is the attention map of the translated image by the attention network $A_T$, and similar for the target domain.

\begin{figure*}
\begin{center}
\includegraphics[width=1.0\linewidth]{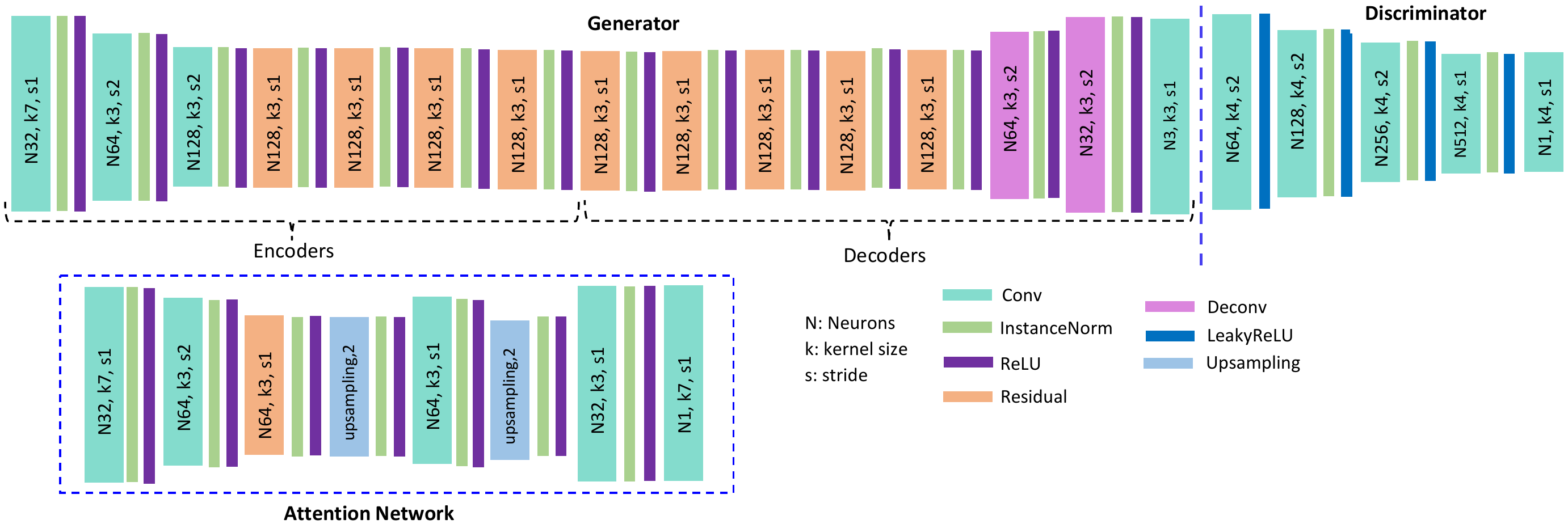}
\end{center}
\caption{Architecture of the generator, discriminator and attention networks where $N$, $k$, and $s$ represent the number of neurons, kernel size and stride respectively.}
\label{fig:GAN}
\end{figure*}

\subsection{Discriminative Re-ID Module}
We integrate the proposed discriminative re-ID module with image generation framework through sharing the encoder as the backbone  for re-ID, allowing it to learn more discriminative features by leveraging shared knowledge. The triplet loss is widely adopted by researchers to boost re-ID performance as it pushes images of the same identity closer to each other than features from dissimilar identities. For example, considering the three images $x_i^1, x_i^2, x_i^3$ which are the anchor, positive and negative image respectively, the triplet loss can be defined as,
%\vspace{-2mm}
\begin{align}
L_{triplet} = \sum_{i=1}^{n} \Big(max\big\{ D(x_i^1, x_i^2) - D (x_i^1, x_i^3),\tau_1 \big\}\Big).
\label{eq:6}
\end{align}
Although great success has been had with the triplet loss for person Re-ID, Khatun et al. \cite{Amena} argue that it suffers from poor generalisation in real-world scenarios due to totally unseen target data as the triplet loss pushes images of the same identity close to each other only when probe images come from the same identity, which is not practical in the real world. Thus, a quartet loss is introduced in \cite{Amena} which improves the traditional triplet loss and can be represented by,
\begin{align}
L_{quartet} = \sum_{i=1}^{n} \Big(max\big\{ D(x_i^1, x_i^2) - D (x_i^1, x_i^3) \nonumber \\+  D(x_i^1, x_i^2) - D (x_i^4, x_i^3),\tau_1 \big\}\Big).
\label{eq:6}
\end{align}
Following \cite{Amena}, the quartet loss has four input images instead of three: an anchor, a positive and two negative images. The quartet loss will then minimise the intra-class variation over the inter-class variation, regardless of whether the probe image belongs to the same person or not, improving the generalisation of the network.

In our proposed EDAAN, we adopt both the triplet and quartet loss and investigate which one is most promising for boosting re-ID performance. These losses force the encoder to minimise the intra-class variation by pulling the feature maps of the same identity together. We further use softmax classification loss to encourage the feature maps of different identities to be separated in feature space. The classification loss is,
\begin{equation}
L_{id}= -log(p(x)),
\label{eq:7}
\end{equation}
where p(.) is the predicted probability that the input image belongs
to the ground-truth class based on its feature map.

The overall objective of our proposed EDAAN thus becomes,
\begin{multline}
L_{total}(\!A_S, \!A_T, \!G, \!F, \!D_S, \!D_T)\!= \!L_{GAN}^S + L_{GAN}^T  + \lambda_{attn} (L_{attn}\\(A_S) + L_{attn}(A_T)) +  \lambda_{quartet}L_{quartet} +\lambda_{id}L_{id}, 
\label{eq:8}
\end{multline}
where $L_{GAN}^S$ and $L_{GAN}^T$ are the adversarial losses for the source and target domain, and $\lambda_{attn}$, $\lambda_{quartet}$ and $\lambda_{id}$ are the weights of the related loss terms. We set $\lambda_{attn}$ = 10, $\lambda_{quartet}$ = 1 and $\lambda_{id}$ = 1 throughout our experiments.

%\begin{multline}
%L_{total}(\!A_S, \!A_T, \!G, \!F, \!D_S, \!D_T)\!= \!L_{GAN}^S + L_{GAN}^T  +\\ \lambda_{cyc} (L_{cyc}^S +L_{cyc}^T )  + \lambda_{attn} (L_{attn}(A_S) + L_{attn}(A_T)) + \\ \lambda_{quartet}L_{quartet} +\lambda_{id}L_{id}, 
%\label{eq:8}
%\end{multline}

\section{Experiments}
\subsection{Dataset and Evaluation Protocol}

In this work, we evaluate our proposed method on four re-ID datasets: Market-1501, DukeMTMC-reID, CUHK03, and PRID2011. 

(1) \textbf{Market1501} \cite{Market} dataset contains 32,643 images of 1,501 identities captured by six cameras in Tsinghua
University. Subjects are automatically detected by the deformable part model (DPM) detector. It consists of 12,936 training images of 751 identities and 19,732 testing images of 750 identities in a close to real-world setting. There is an average of 17.2 images per training identity. \textbf{\textit{Evaluation Protocol:}} We followed evaluation protocol of \cite{Wei2018PersonTG}.

\vspace{0.5mm}
(2) \textbf{DukeMTMC-reID} \cite{Duke} has recently been released for evaluating the efficiency of  multi-target multi-camera tracking systems. It is composed of 36,411 hand-drawn bounding boxes of 1,404 identities that covers a a single outdoor scene extracted from eight high resolution cameras. For the training set, 702 identities are used which comprises 16,522 bounding boxes while the remaining 702 identities are used for testing, where there are 17,661 gallery images and 2,228 query images. \textbf{\textit{Evaluation Protocol:}} We followed the same protocol as \cite{Wei2018PersonTG} where 1,404 identities are equally divided into groups of 702 for training and testing under a single-query setting.

\vspace{0.5mm}
(3) \textbf{CUHK03} \cite{6909421} consists of 14,097 images of 1,467 identities collected from six surveillance cameras in the CUHK campus. There are 28,192 bounding boxes annotated by both a DPM detector and manually. In this dataset, each identity is taken from two disjoint camera views. This dataset is challenging due to background clutter and occlusion.  \textbf{\textit{Evaluation Protocol:}} For a fair comparison, 1,367 identities are used for training, and 100 identities are used for testing following \cite{Wei2018PersonTG}. 

\vspace{0.5mm}
(4) \textbf{PRID2011} \cite{conf/scia/HirzerBRB11} dataset extracts images from video recorded by static surveillance cameras with two camera views, each of which contains 385 and 749 identities. Among the 1,134 persons, only 200 are common to both camera views. \textbf{\textit{Evaluation Protocol:}} For this dataset, we followed the same protocol as \cite{Wei2018PersonTG} where 100 identities are used for testing.

%\vspace{-3mm}
\subsection{ Implementation Details}
We use PyTorch \cite{NEURIPS2019_9015} to train EDAAN and for all experiments the Adam optimizer is used to train from scratch with a batch size of 16 and a learning rate of 0.0002. We train for 200 epochs. The learning rate is constant for the first 100 epochs, and then linearly decays towards zero over the next 100 epochs. The attention networks seek to isolate the background and thus ignore the image foreground, however, the discriminator considers the entire image and not only the attended regions. As such we train the discriminator on whole images for the first 30 epochs, and afterwards train with the masked images only so that it will pay attention only to the attended regions. As the estimation value ranges from 0 to 1 per-pixel, the attention map slowly converges to 1, thus the attention network will slowly pay attention to the foreground alongside the background of an image. To tackle this issue, we train the attention networks for the first 30 epochs only. The architecture of the generator, discriminator and attention networks are illustrated in Figure \ref{fig:GAN}. The classifier uses the feature vector as the input for the classification loss. It contains a 128-dim fully connected layer, batch normalisation, dropout, ReLU and finally the output layer is a fully connected layer with logits equivalent to the number of identities. We set the dropout rate to 0.5. The embedding layer is a fully-connected layer for the quartet loss that maps the feature vector to a 128-dim embedding vector. During the test, the feature learner's re-ID model is used directly for the target dataset.

%DWe use the Generator $G$ to translate the images from the source to the target domain, and the Generator $F$ is utilised to translate the images from the target to source domain during testing. During testing, we employ the Generator $G$ for the source $\rightarrow$ target domain image translation and the Generator $F$ for the target $\rightarrow$ source domain image translation. For the classification loss, the classifier takes the feature vector as the input, and includes a 128-dim fully connected (FC) layer followed by batch normalization, dropout, and ReLU as the middle layer, and an FC layer with logits equal to the number of identities as the output layer. The dropout rate is set at 0.5 empirically. For the quartet loss, the embedding layer is an FC layer that maps the feature vector to a 128-dim embedding vector. 

\begin{table*}
\fontsize{9.0}{9.0}\selectfont
\begin{center}
\caption{\label{tab:result} Cross-domain performance comparison of EDAAN with state-of-the-art methods on target domains. When the model is tested on DukeMTMC-reID, Market-1501 is used as the source domain and vice-versa. R1, R5, and R10 indicates rank-1, rank-5, rank-10 identification accuracy respectively, and mAP is the mean average precision score.}
\begin{tabular}{|p{3.5cm}|p{0.8cm}| p{0.8cm} |p{0.8cm}| p{0.8cm}||p{0.8cm} |p{0.8cm}| p{0.8cm}| p{0.8cm}|}
\hline
\multirow{2}{*}{Method}  & \multicolumn{4}{c|}{Market1501 $\rightarrow$ DukeMTMC-reID} &   \multicolumn{4}{c|}{DukeMTMC-reID $\rightarrow$ Market1501 } \\
\cline{2-9}
%& \multicolumn{2}{c|}{cam1/cam2 } & \multicolumn{2}{c|}{cam2/cam1}  & \multicolumn{2}{c|}{cam1/cam2 } & \multicolumn{2}{c|}{cam2/cam1} \\
\cline{2-9}
& R1  & R5 & R10  & mAP & R1  & R5 & R10  & mAP\\
\hline
%Supervised Learning &66.7 &79.1 &83.8 &46.3 &75.8 &89.6 &92.8 &52.2 & & & & & & & &\\
%Direct Transfer &33.1 &49.3 &55.6 &16.7 &43.1 &60.8 &68.1 &17.0 \\
%Direct Transfer &38.4 &54.3 &61.0 &22.0 &48.1 &66.3 &73.1 &21.2 \\
Direct Transfer &42.4 &56.5 &63.2 &23.0 &52.0 &70.2 &76.5 &22.0 \\
CycleGAN  &44.1 &58.6 &65.0 &23.6 &55.2 &72.8 &79.4 &23.2 \\
%CycleGAN \cite{CycleGAN2017} &38.1 &54.4 &60.5 &19.6 &45.6 &63.8 &71.3 &19.1 \\
%CycleGAN \cite{CycleGAN2017} &40.2 &56.7 &62.8 &22.4 &51.6 &68.1 &75.8 &22.3 \\
PTGAN \cite{Wei2018PersonTG} &27.4 &- &50.7 &- &38.6 &- &66.1 &- \\
%PUL \cite{Fan:2018:UPR:3282485.3243316}&30.0 &43.4 &48.5 &16.4 &45.5 &60.7 &66.7 &20.5 \\
%TJ-AIDL \cite{TJ-AIDL}&44.3 &59.6 &65.0 &23.0 &58.2 &74.8 &81.1 &26.5 \\
%MMFA \cite{MMFA} &45.3 &59.8 &66.3 &24.7 &56.7 &75.0 &81.8 &27.4 \\
SPGAN \cite{image-image18} &41.1 &56.6 &63.0 &22.3 &51.5 &70.1 &76.8 &22.8 \\
ATNet \cite{ATNet}&45.1 &59.5 &64.2 &24.9 &55.7 &73.2 &79.4 &25.6 \\
%SPGAN+LMP \cite{image-image18} &46.9 &62.6 &68.5 &26.4 &58.1 &76.0 &82.7 &26.9 \\
%HHL \cite{HHL}&46.9 &61.0 &66.7 &27.2 &62.2 &78.8 &84.0 &31.4 \\
%BUC \cite{BUC}&47.4 &62.6 &68.4 &27.5 &66.2 &79.6 &84.5 &38.3 \\
%M2M-GAN+LMP \cite{M2M-GAN} &54.4 &- &- &31.6 &63.1 &- &- &30.9 \\
M2M-GAN \cite{M2M-GAN} &49.6 &- &- &26.1 &57.5 &- &- &26.8 \\
%CFSM \cite{CFSM} &49.8 &- &- &27.3 &61.2 &- &- &28.3 \\
CR-GAN \cite{CR-GAN} &52.2 &- &- &30.0 &59.6 &- &- &29.6\\
%TAUDL \cite{TAUDL}&61.7 &- &- &43.5 &63.7 &- &- &41.2 \\
%CAMEL \cite{CAMEL}&- &- &- &- &54.5 &- &- &26.3\\
%ECN \cite{ECN}  &63.3 &75.8 &80.4 &40.4  &75.1 &87.6 &91.6 &43.0\\
\hline
EDAAN with Triplet &55.2 &68.0 &72.6 &33.5 &62.3 &81.8 &84.0 &32.7\\
EDAAN with Quartet &\textbf{57.8} &\textbf{72.2} &\textbf{78.3} &\textbf{39.6} &\textbf{64.5} &\textbf{83.0} &\textbf{86.3} &\textbf{35.4} \\
\hline
\end{tabular}
\end{center}
\end{table*}

\begin{table*}
\fontsize{9.0}{9.0}\selectfont
\begin{center}
\caption{\label{tab:result2} Cross-domain performance comparison of EDAAN with state-of-the-art methods on target domains. When tested on Market-1501 and DukeMTMC-reID datasets, CUHK03 is used as the source domain. R1, R5, and R10 indicates rank-1, rank-5, rank-10 identification accuracy respectively, and mAP is the mean average precision score.}
\begin{tabular}{|p{3.5cm}|p{0.8cm}| p{0.8cm} |p{0.8cm}| p{0.8cm}||p{0.8cm}| p{0.8cm} |p{0.8cm} |p{0.8cm}|}
\hline
\multirow{2}{*}{Method}  &   \multicolumn{4}{c|}{CUHK03 $\rightarrow$ Market1501 }&   \multicolumn{4}{c|}{CUHK03 $\rightarrow$ DukeMTMC-reID }\\
\cline{2-9}
%& \multicolumn{2}{c|}{cam1/cam2 } & \multicolumn{2}{c|}{cam2/cam1}  & \multicolumn{2}{c|}{cam1/cam2 } & \multicolumn{2}{c|}{cam2/cam1} \\
\cline{2-9}
& R1  & R5 & R10  & mAP & R1  & R5 & R10  & mAP \\
\hline
%Supervised Learning &66.7 &79.1 &83.8 &46.3 &75.8 &89.6 &92.8 &52.2 & & & & & & & &\\
%Direct Transfer &30.4 &45.3 &51.6 &19.1 &32.3 &49.7 &55.3 &18.6\\
%CycleGAN \cite{CycleGAN2017}&49.1 &66.4 &72.3 &21.0 &34.7 &50.2 &56.1 &20.5\\
Direct Transfer &48.3 &68.9 &72.2 &19.4 &36.7 &50.5 &56.3 &18.7\\
CycleGAN &50.1 &69.8 &75.4 &21.3 &38.7 &57.7 &59.1 &20.6\\
PTGAN \cite{Wei2018PersonTG}&31.5 &- &60.2 &- &17.6 &- &38.5 &- \\
SPGAN \cite{image-image18} &42.3 &- &- &19.0 &- &- &- &- \\
%HHL \cite{HHL}&56.8 &74.7 &81.4 &29.8 &42.7 &57.5 &64.2 &23.4\\
CR-GAN \cite{CR-GAN} &58.5 &75.8 &81.9 &30.4 &46.5 &61.6 &67.0 &26.9\\
%TAUDL \cite{TAUDL}&63.7 &- &- &41.2 &61.7 &- &- &43.5\\
\hline
%DAAN with Quartet &64.0 &81.6 &86.3 &41.5 &59.7 &80.4 &85.4 &45.1 \\
EDAAN with Triplet &61.4 &78.1 &82.2 &32.5 &48.8 &65.5 &71.0 &32.6\\
EDAAN with Quartet &\textbf{63.0} &\textbf{80.6} &\textbf{85.4} &\textbf{34.5} &\textbf{51.7} &\textbf{70.3} &\textbf{77.8} &\textbf{35.1} \\
\hline
\end{tabular}
\end{center}
\end{table*}

\begin{table*}
\fontsize{9}{9}\selectfont
\begin{center}
\caption{\label{tab:cross} Cross-domain performance comparison on PRID2011 dataset trained with CUHK03 and Market1501 dataset. $cam1/cam2$ indicates that $cam1$ of PRID is used as the query set while $cam2$ is the gallery set and vice-versa. R1 and R10 indicates rank-1 and rank-10 identification accuracy respectively.}
\begin{tabular}{|p{3cm}|p{0.8cm}|p{0.8cm}|p{0.8cm}|p{0.8cm}||p{0.8cm}|p{0.8cm}|p{0.8cm}|p{0.8cm}|}
\hline
\multirow{3}{*}{Method}  & \multicolumn{4}{c|}{CUHK03 $\rightarrow$ PRID} &   \multicolumn{4}{c|}{Market1501 $\rightarrow$ PRID }\\
\cline{2-9}
& \multicolumn{2}{c|}{cam1/cam2 } & \multicolumn{2}{c|}{cam2/cam1}  & \multicolumn{2}{c|}{cam1/cam2 } & \multicolumn{2}{c|}{cam2/cam1} \\
\cline{2-9}
& R1  & R10 & R1  & R10 & R1  & R10 & R1  & R10 \\
\hline
%Direct Transfer &2.0 &11.5 &1.5 &11.5 &5.0 &26.0 &11.0 &40.0\\
Direct Transfer &9.5 &18.5 &7.0 &18.0 &12.0 &33.0 &18.0 &46.0\\
PTGAN(cam1) \cite{Wei2018PersonTG}&18.0   &43.5 &6.5 &24.0 &17.5 &50.5 &8.5 &28.5\\
PTGAN(cam2) \cite{Wei2018PersonTG}&17.5   &53.0 &22.5 &54.0 &10.0 &31.5 &10.5 &37.5\\
ATNet(cam1) \cite{ATNet} &-   &- &- &-  &24.0 &51.5 &21.5 &46.5\\
ATNet(cam2) \cite{ATNet} &-   &- &- &-  &15.0 &51.0 &14.0 &41.5\\
\hline
EDAAN (cam1)  &\textbf{35.0}  &\textbf{52.5} &\textbf{23.6} &\textbf{35.5} &\textbf{34.2} &\textbf{61.5} &\textbf{29.5} &\textbf{54.0}\\
EDAAN (cam2)  &\textbf{32.5}  &\textbf{62.0} &\textbf{41.0} &\textbf{63.0} &\textbf{24.5} &\textbf{58.5} &\textbf{26.0} &\textbf{48.5}\\
\hline
\end{tabular}
\end{center}
\end{table*}
%During testing, the generator, $G_S,E_S$ and $G_T,E_T$ are used for source to target and target to source dataset respectively.

\vspace{-2mm}
\subsection{Comparison with State-of-the-Art Approaches}

We evaluate the performance of the proposed EDAAN in two situations: 1) when the style is transferred from one large domain to another large domain such as from CUHK03 to Market-1501 or DukeMTMC-reID and 2) when the style is transferred from a large domain to a small domain such as from Market-1501, or CUHK03 to PRID2011.

\subsubsection{Performance transferring from large to large domain}
The performance of the proposed method is reported in Table \ref{tab:result} when the styles are transferred between Market1501 and DukeMTMC-reID datasets. We compare our performance with state-of-the-art methods in terms of rank (top-k) accuracy and mean average precision (mAP). \textbf{Direct Transfer} indicates the trained model is directly tested on the target domain without any domain adaptation technique applied. From Table \ref{tab:result}, we can clearly observe that the performance drops severely when the ResNet-50 model is trained on Market-1501 but tested on DukeMTMC-reID, and a similar performance drop is observed when DukeMTMC-reID is used as the source domain. The reason behind these performance degradation is domain shift or dataset bias. However, when Market1501 images are transferred to the DukeMTMC-reID style, the proposed EDAAN model achieves 57.8\% rank-1 accuracy and a  39.6 map score on DukeMTMC-reID dataset, outperforming the previous state-of-the-art method \cite{CR-GAN} by 5.6\% for rank-1 accuracy. We also surpass \cite{CR-GAN} by for  4.9\% rank-1 accuracy when the styles are transferred from DukeMTMC-reID to Market1501 and the model is tested on the Market1501 dataset.

Table \ref{tab:result2} shows the evaluation of the proposed model when styles are transferred from CUHK03 to Market-1501 and DukeMTMC-reID, which constitutes a large domain shift between the source and target domains. The previous state-of-the-art methods \cite{Wei2018PersonTG,CR-GAN} achieve 31.5\% and 58.5\% rank-1 accuracy when tested on Market-1501 dataset, while our method achieves 63.0\% rank-1 accuracy. Similar performance can be seen when tested on DukeMTMC-reID with 5.2\% rank-1 improvement over the state-of-the-art method.

As shown in Table \ref{tab:result} and \ref{tab:result2}, we perform both triplet \cite{Cheng_2016_CVPR} and quartet \cite{Amena} loss for re-ID. In the four considered transfer cases, Market-1501 $\rightarrow$ DukeMTMC-reID, DukeMTMC-reID $\rightarrow$ Market-1501, CUHK03 $\rightarrow$ Market-1501, and CUHK03 $\rightarrow$ DukeMTMC-reID, we observe that the quartet loss \cite{Amena} performs better with the proposed model than the traditional triplet loss.

\begin{table*}
\fontsize{9}{9}\selectfont
\begin{center}
\caption{\label{tab:single} Single-domain performance comparison of EDAAN with state-of-the-art methods on CUHK03, Market-1501, and DukeMTMC-reID datasets under single-domain setting. R1, R5, and R10 indicates rank-1, rank-5, rank-10 identification accuracy respectively, and mAP is the mean average precision score.}
\begin{tabular}{|p{3.0cm}|p{0.6cm}| p{0.6cm} |p{0.6cm}| p{0.6cm}||p{0.6cm} |p{0.6cm}| p{0.6cm}| p{0.6cm}||p{0.6cm} |p{0.6cm}| p{0.6cm}| p{0.6cm}|}
\hline
\multirow{2}{*}{Method} & \multicolumn{4}{c|}{CUHK03} & \multicolumn{4}{c|}{Market1501 } &   \multicolumn{4}{c|}{DukeMTMC-reID } \\
\cline{2-13}
\cline{2-13}
& R1  & R5 & R10  & mAP & R1  & R5 & R10  & mAP & R1  & R5 & R10  & mAP \\
\hline
Supervised Learning &66.7 &79.1 &83.8 &46.3 &75.8 &89.6 &92.8 &52.2 & & & &\\
DCF \cite{Li_2017_CVPR} &74.2 &94.3 &97.6 &-  &80.3 &- &-&- & & & &\\
Spindle Net \cite{zhao2017spindle} & 88.5 &97.8 &98.6 &- &76.9 &91.5 &94.6 &- &- &- &- &-\\
DaRe \cite{DaRe} &73.8 &- &- &74.7 &90.9 &- &-&86.7 &84.4 &- &- &80.0\\
AACN \cite{Xu2018AttentionAwareCN} &91.4 &98.9 &99.5 &- &88.7 &- &-&82.9 &76.8 &- &- &59.3\\
MLFN \cite{98a1e05749b24099a51dcf3c22daefd9} &82.8 &- &- &- &90.0 &- &-&74.3 &81.0 &- &- &62.8\\
CamStyle \cite{zhong2018camera} &- &- &- &- &89.5 &- &-&71.6 &78.3 &- &- &57.6\\
HA-CNN \cite{HA-CNN}&-&-&-&- &91.2&- &- &75.7 &80.5 &- &- &63.8\\
%VPM \cite{VPM} &-&-&-&- &93.0 &97.8 &98.8 &80.8 &83.6 &91.7 &94.2 &72.6\\
AANet \cite{AANet}&-&-&-&-&93.8 &- &98.6 &82.5 &86.4 &- &- &72.6\\
Mancs \cite{Mancs} &93.8 &99.3 &99.8 &-&93.1 &- &- &82.3 &84.9 &- &- &71.8\\
\hline
EDAAN &\textbf{94.7} &\textbf{99.5} &\textbf{99.8} &\textbf{83.4} &\textbf{95.3} &\textbf{97.8} &\textbf{99.6} &\textbf{86.8} &\textbf{88.2} &\textbf{94.5} &\textbf{95.4} &\textbf{83.1} \\
\hline
\end{tabular}
\end{center}
\end{table*}

\begin{table*}
\fontsize{9}{9}\selectfont
\begin{center}
\caption{\label{tab:ablation} Ablation studies on DukeMTMC-reID dataset when the model is trained on Market-1501. ``DAAN w/o attention": without attention and not in end-to-end manner. ``DAAN with attention": with the attention network but without end-to-end. ``EDAAN w/o attention": end-to-end but wihout attention network and ``EDAAN with attention": with attention and image generative and re-ID feature learning module together. R1, R5, and R10 indicates rank-1, rank-5, rank-10 identification accuracy respectively, and mAP is the mean average precision score.}
\begin{tabular}{|p{4.0cm}|p{0.8cm}| p{0.8cm} |p{0.8cm}| p{0.8cm}||p{0.8cm}| p{0.8cm} |p{0.8cm}| p{0.8cm}|}
\hline
\multirow{2}{*}{Method}  &   \multicolumn{4}{c|}{Market1501 $\rightarrow$ DukeMTMC-reID  }  &   \multicolumn{4}{c|}{DukeMTMC-reID $\rightarrow$  Market1501  }\\
\cline{2-9}
\cline{2-9}
& R1  & R5 & R10  & mAP & R1  & R5 & R10  & mAP\\
\hline
DAAN w/o attention module &45.3 &62.5 &67.0 &25.4 &52.2 &73.3 &74.6 &21.2\\
DAAN with attention module&54.2 &70.1 &76.7 &38.1 &61.4 &81.0 &83.5 &33.1\\
EDAAN w/o attention module&48.5 &62.6 &68.4 &26.8 &55.4 &75.6 &77.0 &23.7\\
EDAAN with attention module&\textbf{57.8} &\textbf{72.2} &\textbf{78.3} &\textbf{39.6} &\textbf{64.5} &\textbf{83.0} &\textbf{86.3} &\textbf{35.4}\\
\hline
\end{tabular}
\end{center}
\end{table*} 

\subsubsection{Performance transferring from large to small domain}
To evaluate how the proposed model performs when the styles are transferred from a large to a small domain, we transferred the background styles from CUHK03 to PRID2011 and Market1501 to PRID2011, and the proposed model is then evaluated on PRID2011 as summarised in Table \ref{tab:cross}. From Table \ref{tab:cross}, the model is trained with images from CUHK03 transferred by EDAAN alongside real images, and achieves significant performance gains when tested on PRID, e.g., a 17\% and 14.5\% increase in rank-1 accuracy compared to \cite{Wei2018PersonTG}. Similar improvements can be observed when trained with the transferred images of Martket1501 dataset to other domains, outperforming state-of-the-art approaches \cite{Wei2018PersonTG,ATNet}. 

\begin{figure*}
\begin{center}
\includegraphics[width=1.0\linewidth]{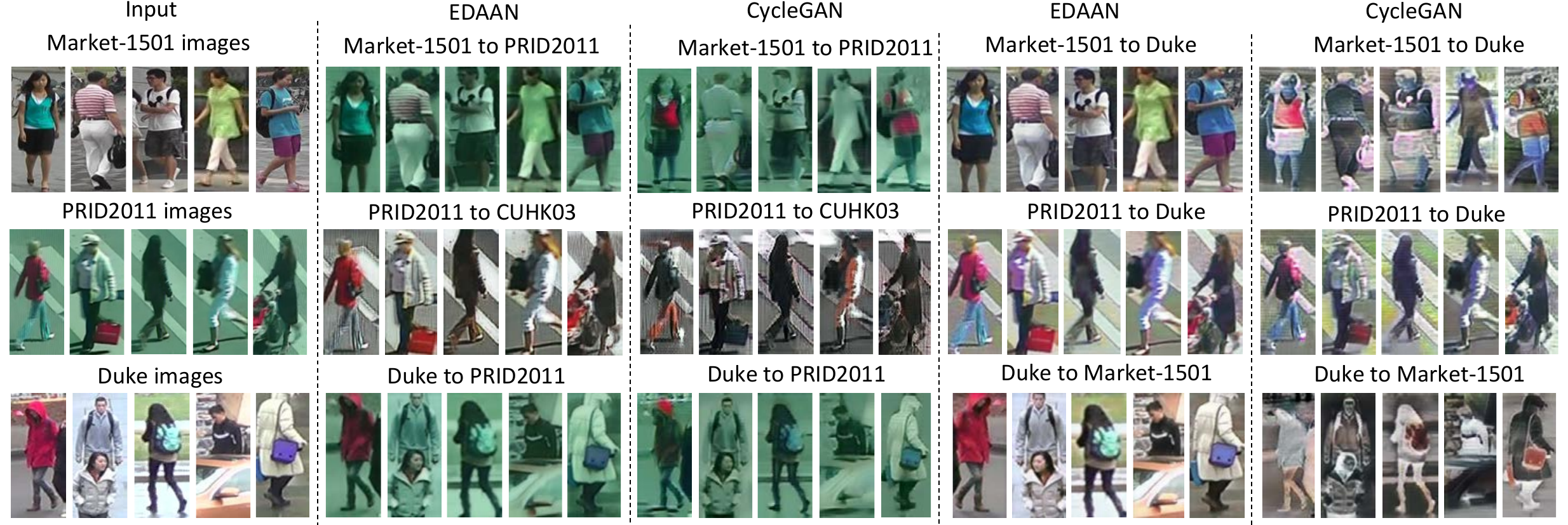}
\end{center}
\vspace{-4mm}
\caption{Examples of the style-transferred images in four domains. The first column represents input images from Market-1501, PRID2011, and Duke datasets. The second and third column shows the transferred images from Market-1501 to PRID2011, PRID2011 to CUHK03, Duke to PRID2011 for the proposed EDAAN and CycleGAN for comparison. Similarly, the fourth and fifth column shows the style transferred images from Market1501 to Duke, PRID2011 to Duke, Duke to Market-1501 for EDAAN and CycleGAN.}
\label{fig:image_comparison}
\end{figure*}

These evaluations on the proposed method collectively indicate the potential of EDAAN for cross-domain learning. Inspecting Figure \ref{fig:image_comparison}, the proposed
method helps preserve the identity and semantic features by preserving the foreground during image domain translation. We can see that, when image styles are transferred from Market-1501 to DukeMTMC-reID, CycleGAN substantially affects the foreground of a person and results in incorrect color in the foreground due to CycleGAN not having an attention mechanism, which hurts person re-ID as we aim to alter the background to match the other domain while preserving the foreground of a person.

\begin{figure*}
\begin{center}
\includegraphics[width=1.0\linewidth]{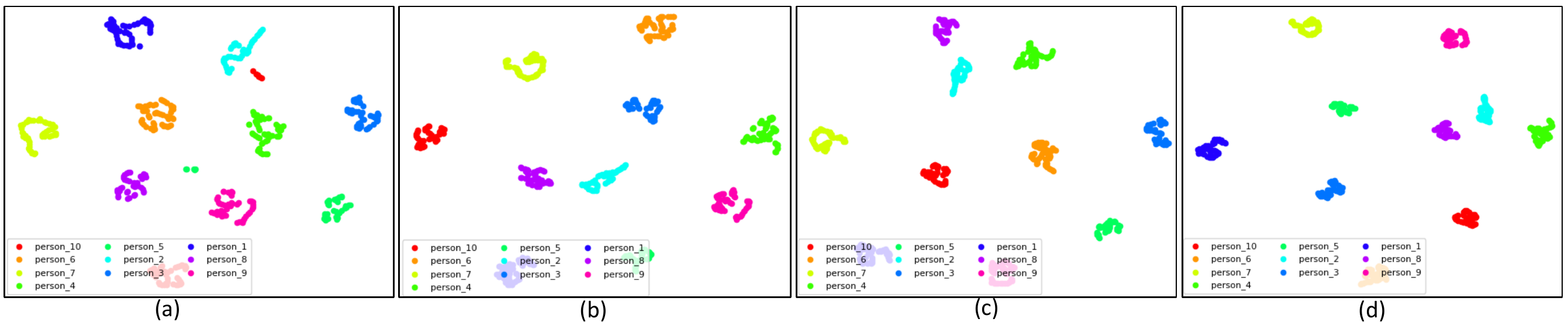}
\end{center}
\caption{t-SNE visualizations of the CNN activations for Market1501 (a) DAAN without attention, (b) EDAAN without attention, (c) DAAN with attention and (d) EDAAN. The 10 different colours correspond 10 different identities. It can be seen that EDAAN has
the tightest grouping of points, indicating that it is best suited to separating the classes.}
\label{fig:tSNE}
\end{figure*}

\subsection{Performance on Single-Domain}
%Our proposed attention-based domain adaptive method transfers the background of a person from the source domain to target domain while preserving the foreground of a person with the help of an attention network, and generates synthetic images with different styles. As EDAAN is trained in an end-to-end fashion, it boosts the performance of person re-ID as the re-ID module is now able to learn more discriminative features as the network is trained with both the real and newly generated images with different styles. 

We also compare the proposed method with state-of-the-art approaches on CUHK03, Market1501, and DukeMTMC-reID under a single domain setting as shown in Table \ref{tab:single}. Our model surpasses previous state-of-the-art methods \cite{Mancs,Xu2018AttentionAwareCN} on CUHK03 by 0.9\% and 3.3\% rank-1 accuracy, respectively. We achieve 95.3\% and 88.2\% rank-1 accuracy on Market-1501 and DukeMTMC-reID, respectively, outperforming previous state-of-the-art methods.

\subsection{Ablation Studies}
%To demonstrate the effectiveness and contribution of each component of the EDAAN, we conduct a series of ablation experiments transferring between DukeMTMC-reID and Market1501 and vice versa. 

We perform a set of ablation experiments to demonstrate the efficacy and contribution of each component of the EDAAN in the context of transferring between DukeMTMC-reID and Market1501. We conduct experiments to verify the influence of the proposed attention module and end-to-end network for person re-ID. To evaluate the effectiveness of the proposed end-to-end network, we conduct experiments on DAAN which consists of two networks: one for image domain translation; and one for re-ID. Thus in the first stage, the proposed attention module and the generative module produces the new synthetic images where the background styles are transferred from the source to target domain. These newly generated style transferred images are then used as the input alongside the real images and fed into another CNN for person re-ID with the quartet loss for verification, and softmax loss for identification.

\textbf{The effect of the Attention module}

To  demonstrate  the  effectiveness  of the proposed attention module, we conduct a series of experiments as shown in Table \ref{tab:ablation} which demonstrate the performance of DAAN w/o attention (i.e. without the attention network), DAAN with attention, EDAAN w/o attention, and the complete approach on DukeMTMC-reID and market-1501 datasets. It can be seen that DAAN w/o the attention network records an 8.9\% drop in rank-1 performance compared to DAAN with the attention network when the model is tested on DukeMTMC-reID dataset. When the model is tested on Market1501 dataset, DAAN achieves 61.4\% rank1 accuracy which is a 9.2\% gain compared to DAAN without the attention network. Similar performance can be observed between EDAAN with and without the attention network. The generated masks from our attention network are shown in Figure \ref{fig:mask}. We also qualitatively compare our method with CycleGAN, as illustrated in Figure \ref{fig:image_comparison}. 

\begin{figure*}
\begin{center}
\includegraphics[width=1.0\linewidth]{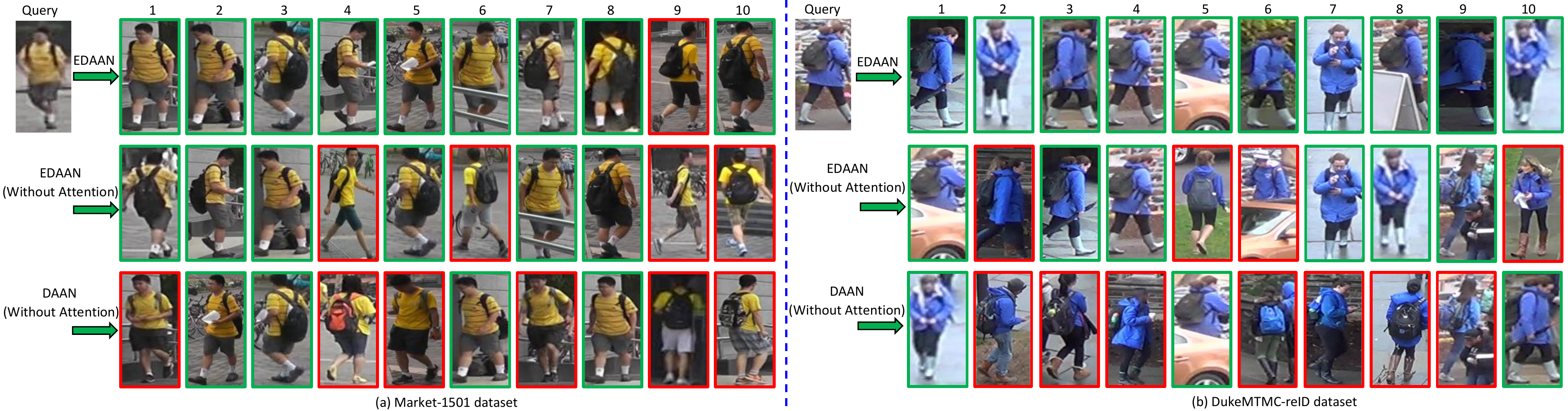}
\end{center}
\caption{Ranking results on the Market-1501 and DukeMTMC-reID datasets for EDAAN, EDAAN (without attention) and DAAN (without attention), where the left-most images are the probe images and images with a green
rectangle indicate the matched person from the gallery sets, and red rectangles indicate a false match. Best viewed in colour.}
\label{fig:Ranking}
\end{figure*}

\begin{figure}
\begin{center}
\includegraphics[width=1.0\linewidth]{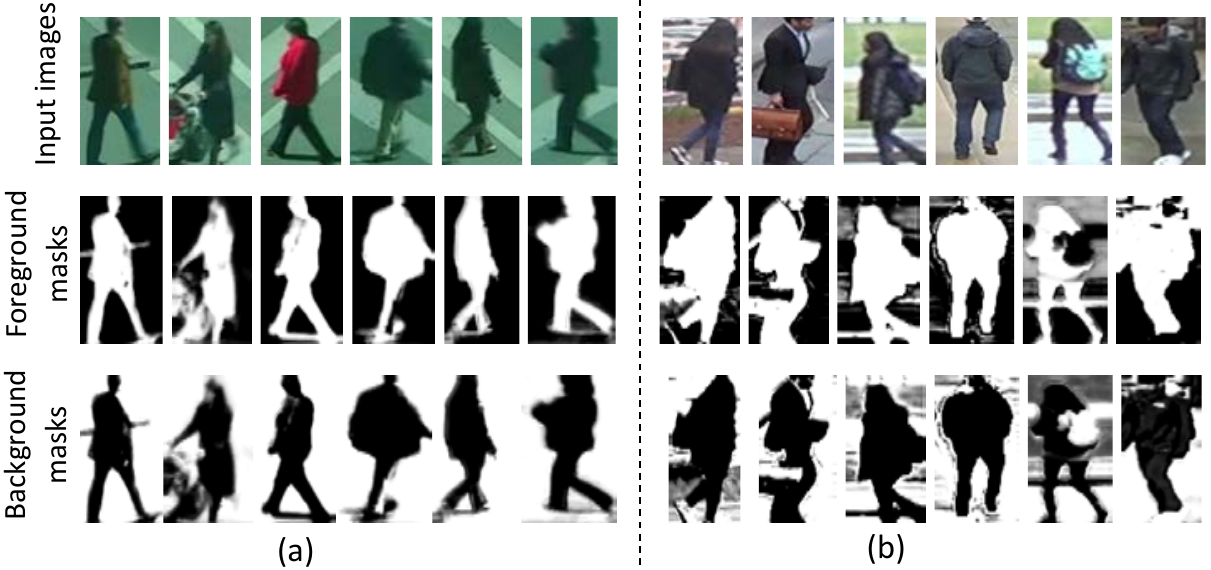}
\end{center}
\caption{Foreground and background attention masks automatically generated by our attention network for (a) PRID2011 and (b) DukeMTMC-reID datasets.}
\label{fig:mask}
\end{figure}

We notice that CycleGAN cannot preserve the consistency of the foreground in most cases. For example, when the Market-1501 images are transferred to the style of PRID2011, CycleGAN changes the color of the tops from blue to red in the first and last images and from green to gray in the fourth image. Similarly, transferring from DukeMTMC-reID to Market1501, CycleGAN failed to preserve the foreground in each image which hurts person re-ID performance. Our attention-based image translation network, by contrast, significantly improves the quality of the translated images and preserves consistency in the foreground while adapting the background from other domains.

\textbf{The effect of an End-to-End network}

EDAAN (end-to-end DAAN) increases the rank-1 accuracy by 3.6\% over DAAN. It is noted that DAAN splits the process into two disjoint steps, limiting the ability of each component to learn from the other. In contrast, the proposed EDAAN incorporates image translation and re-ID feature learning into a single framework, allowing the image generation network to learn discriminative features of a person from the re-ID feature learning module, while the re-ID module learns from the image generation network what the target person looks like. This knowledge sharing strengthens the overall method.

To obtain more insight into the attention and end-to-end network, a t-SNE \cite{tSNE} visualisation is performed on the learned embeddings for the Market1501 dataset as shown in Figure \ref{fig:tSNE}, where 10 classes have been included for better visualization. From Figure \ref{fig:tSNE}, It can be seen that the proposed EDAAN with the attention module optimizes the embedding space such that the data points with the same identity are closer to each other compare to EDAAN without attention and DAAN. It can also be seen that DAAN without attention incorrectly classifies some people, as we can see in Figure \ref{fig:tSNE}(a) the data points of the person 10 are far away from each other in the embedding space. In addition, we show some retrieval examples on Market-1501 and  DukeMTMC-reID  in Figure \ref{fig:Ranking}. The left-most images are the probe images, and in the images to the right of the probe image, the green rectangle denotes a matched image from the top 10 gallery images (best viewed in colour). We compare the ranking results of EDDAN, EDAAN without attention and DAAN without attention, in the first, second, and third rows respectively. From the ranking results, it can be seen that the proposed EDAAN with the attention mechanism can find more correct matches than other methods in the top ranks.

\section{Conclusion}
This paper focuses on addressing the domain shift challenge in person re-ID. Domain shift causes severe performance degradation when trained and tested on different person re-ID datasets. We have proposed a novel technique to jointly translate images between domains and learn discriminative re-ID features in a single framework. To boost the performance on the target domain, we propose a domain adaptive attention network which preserves a person's identity with the aid of the proposed attention network during image domain translation. Specifically, the proposed method can effectively adapt the background from one domain to another without incorrectly modifying the foreground of a person which is crucial for person re-ID. We formulate the network in an end-to-end trainable manner, thus the attention based image translation module leverages discriminative knowledge from the re-ID module and vice-versa, improving the performance of both components. The proposed joint learning network results in a significant performance improvement over state-of-the-art methods on several benchmark datasets.

\if{false}
\section*{Acknowledgment}
The authors would like to thank...
\fi

\ifCLASSOPTIONcaptionsoff
  \newpage
\fi


\bibliographystyle{IEEEtran}
% **** IEEE Bibliography STUFF ****
\bibliography{IEEEabrv,references}

\begin{thebibliography}{10}
\providecommand{\url}[1]{#1}
\csname url@samestyle\endcsname
\providecommand{\newblock}{\relax}
\providecommand{\bibinfo}[2]{#2}
\providecommand{\BIBentrySTDinterwordspacing}{\spaceskip=0pt\relax}
\providecommand{\BIBentryALTinterwordstretchfactor}{4}
\providecommand{\BIBentryALTinterwordspacing}{\spaceskip=\fontdimen2\font plus
\BIBentryALTinterwordstretchfactor\fontdimen3\font minus
  \fontdimen4\font\relax}
\providecommand{\BIBforeignlanguage}[2]{{%
\expandafter\ifx\csname l@#1\endcsname\relax
\typeout{** WARNING: IEEEtran.bst: No hyphenation pattern has been}%
\typeout{** loaded for the language `#1'. Using the pattern for}%
\typeout{** the default language instead.}%
\else
\language=\csname l@#1\endcsname
\fi
#2}}
\providecommand{\BIBdecl}{\relax}
\BIBdecl

\bibitem{XQDA}
S.~Liao, Y.~Hu, X.~Zhu, and S.~Z. Li, ``Person re-identification by local
  maximal occurrence representation and metric learning,'' in \emph{CVPR},
  2015.

\bibitem{KISSME}
M.~Koestinger, M.~Hirzer, P.~Wohlhart, P.~M. Roth, and H.~Bischof, ``Large
  scale metric learning from equivalence constraints,'' in \emph{CVPR}, 2012.

\bibitem{Zheng2016PersonRP}
L.~Zheng, Y.~Yang, and A.~G. Hauptmann, ``Person re-identification: Past,
  present and future,'' \emph{ArXiv}, vol. abs/1610.02984, 2016.

\bibitem{DBLP:journals/corr/VariorHW16}
R.~R. Varior, M.~Haloi, and G.~Wang, ``Gated siamese convolutional neural
  network architecture for human re-identification,'' in \emph{ECCV}, 2016.

\bibitem{Cheng_2016_CVPR}
D.~Cheng, Y.~Gong, S.~Zhou, J.~Wang, and N.~Zheng, ``Person re-identification
  by multi-channel parts-based cnn with improved triplet loss function,'' in
  \emph{CVPR}, 2016.

\bibitem{coral}
B.~Sun, J.~Feng, and K.~Saenko, ``Return of frustratingly easy domain
  adaptation,'' in \emph{AAAI}, 2016.

\bibitem{dcoral}
B.~Sun and K.~Saenko, ``Deep coral: Correlation alignment for deep domain
  adaptation,'' in \emph{ECCV Workshops}, 2016.

\bibitem{G2A}
S.~Sankaranarayanan, Y.~Balaji, C.~D. Castillo, and R.~Chellappa, ``Generate to
  adapt: Aligning domains using generative adversarial networks,'' in
  \emph{CVPR}, 2018.

\bibitem{DupGAN}
L.~Hu, M.~Kan, S.~Shan, and X.~Chen, ``Duplex generative adversarial network
  for unsupervised domain adaptation,'' in \emph{CVPR}, 2018.

\bibitem{CycleGAN2017}
J.-Y. Zhu, T.~Park, P.~Isola, and A.~A. Efros, ``Unpaired image-to-image
  translation using cycle-consistent adversarial networks,'' in \emph{ICCV},
  2017.

\bibitem{8579014}
Y.~{Choi}, M.~{Choi}, M.~{Kim}, J.~{Ha}, S.~{Kim}, and J.~{Choo}, ``Stargan:
  Unified generative adversarial networks for multi-domain image-to-image
  translation,'' in \emph{CVPR}, 2018.

\bibitem{Wei2018PersonTG}
L.~Wei, S.~Zhang, W.~Gao, and Q.~Tian, ``Person transfer gan to bridge domain
  gap for person re-identification,'' in \emph{CVPR}, 2018.

\bibitem{Bak_2018_ECCV}
S.~Bak, P.~Carr, and J.-F. Lalonde, ``Domain adaptation through synthesis for
  unsupervised person re-identification,'' in \emph{ECCV}, 2018.

\bibitem{image-image18}
W.~Deng, L.~Zheng, Q.~Ye, G.~Kang, Y.~Yang, and J.~Jiao, ``Image-image domain
  adaptation with preserved self-similarity and domain-dissimilarity for person
  re-identification,'' in \emph{CVPR}, 2018.

\bibitem{zhong2018camera}
Z.~Zhong, L.~Zheng, Z.~Zheng, S.~Li, and Y.~Yang, ``Camera style adaptation for
  person re-identification,'' in \emph{CVPR}, 2018.

\bibitem{Amena}
A.~Khatun, S.~Denman, S.~Sridharan, and C.~Fookes, ``A deep four-stream siamese
  convolutional neural network with joint verification and identification loss
  for person re-detection,'' in \emph{WACV}, 2018.

\bibitem{AAAI1714313}
W.~Chen, X.~Chen, J.~Zhang, and K.~Huang, ``A multi-task deep network for
  person re-identification,'' in \emph{AAAI}, 2017.

\bibitem{6909421}
W.~Li, R.~Zhao, T.~Xiao, and X.~Wang, ``Deepreid: Deep filter pairing neural
  network for person re-identification,'' in \emph{CVPR}, 2014.

\bibitem{7299016}
E.~Ahmed, M.~Jones, and T.~K. Marks, ``An improved deep learning architecture
  for person re-identification,'' in \emph{CVPR}, 2015.

\bibitem{7780513}
F.~Wang, W.~Zuo, L.~Lin, D.~Zhang, and L.~Zhang, ``Joint learning of
  single-image and cross-image representations for person re-identification,''
  in \emph{CVPR}, 2016.

\bibitem{DBLP:journals/corr/VariorSLXW16}
R.~R. Varior, B.~Shuai, J.~Lu, D.~Xu, and G.~Wang, ``A siamese long short-term
  memory architecture for human re-identification,'' in \emph{ECCV}, 2016.

\bibitem{6976727}
D.~Yi, Z.~Lei, S.~Liao, and S.~Z. Li, ``Deep metric learning for person
  re-identification,'' in \emph{ICPR}, 2014.

\bibitem{6909576}
J.~{Wang}, Y.~{Song}, T.~{Leung}, C.~{Rosenberg}, J.~{Wang}, J.~{Philbin},
  B.~{Chen}, and Y.~{Wu}, ``Learning fine-grained image similarity with deep
  ranking,'' in \emph{CVPR}, 2014.

\bibitem{Ding:2015:DFL:2796563.2796623}
S.~Ding, L.~Lin, G.~Wang, and H.~Chao, ``Deep feature learning with relative
  distance comparison for person re-identification,'' \emph{Pattern
  Recognition}, 2015.

\bibitem{DBLP:journals/corr/ChenCZH17}
W.~Chen, X.~Chen, J.~Zhang, and K.~Huang, ``Beyond triplet loss: a deep
  quadruplet network for person re-identification,'' in \emph{CVPR}, 2017.

\bibitem{8099845}
J.~Lin, L.~Ren, J.~Lu, J.~Feng, and J.~Zhou, ``Consistent-aware deep learning
  for person re-identification in a camera network,'' in \emph{CVPR}, 2017.

\bibitem{Zhao_2017_ICCV}
L.~Zhao, X.~Li, Y.~Zhuang, and J.~Wang, ``Deeply-learned part-aligned
  representations for person re-identification,'' in \emph{ICCV}, 2017.

\bibitem{Zhang2017AlignedReIDSH}
X.~Zhang, H.~Luo, X.~Fan, W.~Xiang, Y.~Sun, Q.~Xiao, W.~Jiang, C.~Zhang, and
  J.~Sun, ``Alignedreid: Surpassing human-level performance in person
  re-identification,'' \emph{CoRR}, 2017.

\bibitem{98a1e05749b24099a51dcf3c22daefd9}
X.~Chang, T.~M. Hospedales, and T.~Xiang, ``Multi-level factorisation net for
  person re-identification,'' in \emph{CVPR}, 2018.

\bibitem{Li_2017_CVPR}
D.~Li, X.~Chen, Z.~Zhang, and K.~Huang, ``Learning deep context-aware features
  over body and latent parts for person re-identification,'' in \emph{CVPR},
  2017.

\bibitem{Yang2019PatchBasedDF}
Q.~Yang, H.-X. Yu, A.~Wu, and W.-S. Zheng, ``Patch-based discriminative feature
  learning for unsupervised person re-identification,'' in \emph{CVPR}, 2019.

\bibitem{zhao2017spindle}
H.~Zhao, M.~Tian, S.~Sun, J.~Shao, J.~Yan, S.~Yi, X.~Wang, and X.~Tang,
  ``Spindle net: Person re-identification with human body region guided feature
  decomposition and fusion.''\hskip 1em plus 0.5em minus 0.4em\relax CVPR,
  2017.

\bibitem{Fan:2018:UPR:3282485.3243316}
H.~Fan, L.~Zheng, C.~Yan, and Y.~Yang, ``Unsupervised person re-identification:
  Clustering and fine-tuning,'' \emph{ACM Trans. Multimedia Comput. Commun.
  Appl.}, 2018.

\bibitem{Lin2019ABC}
Y.~Lin, X.~Dong, L.~Zheng, Y.~Yan, and Y.~Yang, ``A bottom-up clustering
  approach to unsupervised person re-identification,'' in \emph{AAAI}, 2019.

\bibitem{DECAMEL}
H.-X. Yu, A.~Wu, and W.-S. Zheng, ``Unsupervised person re-identification by
  deep asymmetric metric embedding,'' \emph{TPAMI}, 2019.

\bibitem{8976262}
M.~{Ye} and P.~C. {Yuen}, ``Purifynet: A robust person re-identification model
  with noisy labels,'' \emph{TIFS}, 2020.

\bibitem{8922622}
K.~L. {Navaneet}, R.~K. {Sarvadevabhatla}, S.~{Shekhar}, R.~{Venkatesh Babu},
  and A.~{Chakraborty}, ``Operator-in-the-loop deep sequential multi-camera
  feature fusion for person re-identification,'' \emph{TIFS}, 2020.

\bibitem{8732420}
M.~{Ye}, X.~{Lan}, Z.~{Wang}, and P.~C. {Yuen}, ``Bi-directional
  center-constrained top-ranking for visible thermal person
  re-identification,'' \emph{TIFS}, 2020.

\bibitem{Khatun_2020_WACV}
A.~Khatun, S.~DENMAN, S.~Sridharan, and C.~Fookes, ``Semantic consistency and
  identity mapping multi-component generative adversarial network for person
  re-identification,'' in \emph{WACV}, 2020.

\bibitem{AlignGAN}
G.~Wang, T.~Zhang, J.~Cheng, S.~Liu, Y.~Yang, and Z.~Hou, ``Rgb-infrared
  cross-modality person re-identification via joint pixel and feature
  alignment,'' in \emph{ICCV}, 2019.

\bibitem{DBLP:conf/icml/LongC0J15}
M.~Long, Y.~Cao, J.~Wang, and M.~I. Jordan, ``Learning transferable features
  with deep adaptation networks,'' in \emph{ICML}, 2015.

\bibitem{sun2016deep}
B.~Sun and K.~Saenko, ``Deep coral: Correlation alignment for deep domain
  adaptation,'' in \emph{ECCV}, 2016.

\bibitem{RAHMAN2019107124}
M.~M. Rahman, C.~Fookes, M.~Baktashmotlagh, and S.~Sridharan,
  ``Correlation-aware adversarial domain adaptation and generalization,''
  \emph{Pattern Recognition}, 2019.

\bibitem{Li_2018_CVPR_Workshops}
Y.-J. Li, F.-E. Yang, Y.-C. Liu, Y.-Y. Yeh, X.~Du, and Y.-C. Frank~Wang,
  ``Adaptation and re-identification network: An unsupervised deep transfer
  learning approach to person re-identification,'' in \emph{CVPR Workshops},
  2018.

\bibitem{mnih2014recurrent}
V.~Mnih, N.~Heess, A.~Graves \emph{et~al.}, ``Recurrent models of visual
  attention,'' in \emph{NIPS}, 2014.

\bibitem{zhang2018self}
H.~Zhang, I.~Goodfellow, D.~Metaxas, and A.~Odena, ``Self-attention generative
  adversarial networks,'' \emph{arXiv preprint arXiv:1805.08318}, 2018.

\bibitem{yang2017lr}
J.~Yang, A.~Kannan, D.~Batra, and D.~Parikh, ``Lr-gan: Layered recursive
  generative adversarial networks for image generation,'' \emph{arXiv preprint
  arXiv:1703.01560}, 2017.

\bibitem{NIPS2016_6194}
C.~Vondrick, H.~Pirsiavash, and A.~Torralba, ``Generating videos with scene
  dynamics,'' in \emph{NIPS}, 2016.

\bibitem{chen2018attention}
X.~Chen, C.~Xu, X.~Yang, and D.~Tao, ``Attention-gan for object transfiguration
  in wild images,'' in \emph{ECCV}, 2018.

\bibitem{Market}
L.~Zheng, L.~Shen, L.~Tian, S.~Wang, J.~Wang, and Q.~Tian, ``Scalable person
  re-identification: A benchmark,'' in \emph{ICCV}, 2015.

\bibitem{Duke}
E.~Ristani, F.~Solera, R.~S. Zou, R.~Cucchiara, and C.~Tomasi, ``Performance
  measures and a data set for multi-target, multi-camera tracking,'' in
  \emph{ECCV Workshops}, 2016.

\bibitem{conf/scia/HirzerBRB11}
M.~Hirzer, C.~Beleznai, P.~M. Roth, and H.~Bischof, ``Person re-identification
  by descriptive and discriminative classification.'' in \emph{SCIA}, 2011.

\bibitem{NEURIPS2019_9015}
A.~Paszke, S.~Gross, F.~Massa, A.~Lerer, J.~Bradbury, G.~Chanan, T.~Killeen,
  Z.~Lin, N.~Gimelshein, L.~Antiga, A.~Desmaison, A.~Kopf, E.~Yang, Z.~DeVito,
  M.~Raison, A.~Tejani, S.~Chilamkurthy, B.~Steiner, L.~Fang, J.~Bai, and
  S.~Chintala, ``Pytorch: An imperative style, high-performance deep learning
  library,'' in \emph{NIPS}, 2019.

\bibitem{ATNet}
J.~Liu, Z.-J. Zha, D.~Chen, R.~Hong, and M.~Wang, ``Adaptive transfer network
  for cross-domain person re-identification,'' in \emph{CVPR}, 2019.

\bibitem{M2M-GAN}
W.~Liang, G.~Wang, J.~Lai, and J.~Z. Zhu, ``M2m-gan: Many-to-many generative
  adversarial transfer learning for person re-identification,'' \emph{AAAI},
  2019.

\bibitem{CR-GAN}
Y.~Chen, X.~Zhu, and S.~Gong, ``Instance-guided context rendering for
  cross-domain person re-identification,'' in \emph{ICCV}, 2019.

\bibitem{DaRe}
Y.~Wang, L.~Wang, Y.~You, X.~U. Zou, V.~Chen, S.~P. Li, G.~Huang, B.~Hariharan,
  and K.~Q. Weinberger, ``Resource aware person re-identification across
  multiple resolutions,'' in \emph{CVPR}, 2018.

\bibitem{Xu2018AttentionAwareCN}
J.~Xu, R.~Zhao, F.~Zhu, H.~Wang, and W.~Ouyang, ``Attention-aware compositional
  network for person re-identification,'' in \emph{CVPR}, 2018.

\bibitem{HA-CNN}
W.~Li, X.~Zhu, and S.~Gong, ``Harmonious attention network for person
  re-identification,'' in \emph{CVPR}, 2018.

\bibitem{AANet}
C.-P. Tay, S.~Roy, and K.-H. Yap, ``Aanet: Attribute attention network for
  person re-identifications,'' in \emph{CVPR}, 2019.

\bibitem{Mancs}
C.~Wang, Q.~Zhang, C.~Huang, W.~Liu, and X.~Wang, ``Mancs: A multi-task
  attentional network with curriculum sampling for person re-identification,''
  in \emph{ECCV}, 2018.

\bibitem{tSNE}
L.~Van Der~Maaten, ``Accelerating t-sne using tree-based algorithms,''
  \emph{JMLR}, 2014.

\end{thebibliography}
%LPR: i.e. references.bib in this folder is your references database.
%LPR: I use JabRef to manage my references.

% Only show authors if fmtNOAUTH is false
\ifLPRfmtNOAUTH\else
%\newpage
\vspace{-0.5 cm}

%\begin{IEEEbiography}[{\includegraphics[width=1in,height=1.25in,clip,keepaspectratio]{photos/renaud.jpg}}]{Luke Renaud}
%\begin{IEEEbiographynophoto}{Luke Renaud}
%(S’13) received the B.S. in Electrical Engineering \textit{summa cumme laude} with a from Washington State University in Pullman, WA in 2013. He is currently pursuing a Ph.D. in RF Microelectronics from Washington State University, Pullman, WA.

%If you're reading this paragraph as a reviewer, then the person who submit the paper has forgotten to remove my template BIO from their document. You really should reject their paper, that's pretty sloppy.
%\vspace{-0.8 cm}
%\end{IEEEbiography}

%\vfill

%\begin{IEEEbiography}[{\includegraphics[width=1in,height=1.25in,clip,keepaspectratio]{photos/ylabmate}}]{Your Labmate}
%\begin{IEEEbiographynophoto}{Your Labmate}
%Bio paragraph
%\end{IEEEbiography}
%\vspace{-0.8 cm}

%\begin{IEEEbiography}[{\includegraphics[width=1in,height=1.25in,clip,keepaspectratio]{photos/yprof}}]{Your Professor}
%\begin{IEEEbiographynophoto}{Your Professor}
%Bio paragraph
%\end{IEEEbiography}

%\newpage



\vfill


\fi

% Add the random todo items to the end of the document if we want them.
\ifLPRfmtDRAFT
\fi

\end{document}